\documentclass[letterpaper, 10 pt, conference]{ieeeconf}  

\IEEEoverridecommandlockouts                              

\usepackage{cite}
\usepackage{amsmath,amssymb,amsfonts}
\usepackage{algorithmic}
\usepackage{graphicx}
\usepackage{textcomp}
\usepackage{xcolor}

\usepackage{subcaption}
\usepackage{booktabs}
\usepackage{mathtools}

\usepackage{textcomp} 
\usepackage{siunitx}
\usepackage{adjustbox}
\usepackage{pgfplots}

\usepackage{pgf}
\usepackage{multirow}
\usepackage{hyperref}
	
\usepackage{wrapfig}
\usepackage{tablefootnote}
\usepackage{placeins}
\usepackage{tikz}

\newcommand{\IDOOD}{\ensuremath{\overline{\text{ID+OOD}}}}

\usepackage{pifont}
\newcommand{\xmark}{\ding{55}}%

\newcommand\copyrighttext{%
	\footnotesize \textcopyright 2025 IEEE. Personal use of this material is permitted.  Permission from IEEE must be obtained for all other uses, in any current or future media, including reprinting/republishing this material for advertising or promotional purposes, creating new collective works, for resale or redistribution to servers or lists, or reuse of any copyrighted component of this work in other works.}
\newcommand\copyrightnotice{%
	\begin{tikzpicture}[remember picture,overlay]
	\node[anchor=south,xshift=5pt,yshift=10pt] at (current page.south) {\fbox{\parbox{\dimexpr\textwidth-\fboxsep-\fboxrule\relax}{\copyrighttext}}};
	\end{tikzpicture}%
}

\title{\LARGE \bf
LoRD: Adapting Differentiable Driving Policies to Distribution Shifts
}

\author{Christopher Diehl$^{1}$, Peter Karkus$^{2}$, Sushant Veer$^{2}$, Marco Pavone$^{2,3}$, and Torsten Bertram$^{1}$
\thanks{$^{1}$Institute of Control Theory and
Systems Engineering, TU Dortmund University. $^{2}$ NVIDIA Research. $^{3}$  Department of Aeronautics and Astronautics, Stanford University. Contact: \texttt{forename.surname@tu-dortmund.de}, \texttt{\{sveer,pkarkus,mpavone\}@nvidia.com} }%
}

\begin{document}

\maketitle
\thispagestyle{empty}
\pagestyle{empty}

\selectfont

\copyrightnotice

\begin{abstract}
    Distribution shifts between operational domains can severely affect the performance of learned models in self-driving vehicles (SDVs). While this is a well-established problem, prior work has mostly explored naive solutions such as fine-tuning, focusing on the motion prediction task. In this work, we explore novel adaptation strategies for differentiable autonomy stacks (structured policy) consisting of prediction, planning, and control, perform evaluation in closed-loop, and investigate the often-overlooked issue of catastrophic forgetting. Specifically, we introduce two simple yet effective techniques: a low-rank residual decoder (LoRD) and multi-task fine-tuning.
    Through experiments across three models conducted on two real-world autonomous driving datasets (nuPlan, exiD), we demonstrate the effectiveness of our methods and highlight a significant performance gap between open-loop and closed-loop evaluation in prior approaches. Our approach improves forgetting by up to 23.33\% and the closed-loop out-of-distribution driving score by 9.93\% in comparison to standard fine-tuning. \\ \url{https://github.com/rst-tu-dortmund/LoRD}{}
\end{abstract}

\section{Introduction}

\begin{figure}[t]
    \centering
    \includegraphics{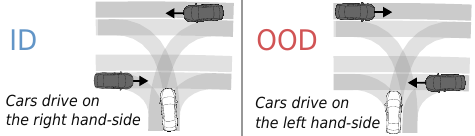}
    \caption{SDVs need to adapt to various distribution shifts, such as traffic regulations, social norms, traffic density, and weather conditions.}
    \label{fig::intro}
    \vspace{-0.5cm}
\end{figure}

Machine learning-based SDV architectures have demonstrated impressive performance across many tasks. However, they often struggle with distribution shifts between training and deployment scenarios. As visualized in the SDV example in Fig \ref{fig::intro}, distribution shifts can result from differences in traffic rules, social norms, weather conditions, traffic density, vehicle types, etc., and can cause severe degradation in model accuracy, leading to catastrophic failures \cite{deHaanNeurIPS2019}. For instance, a recent study on trajectory prediction \cite{UniTraj} revealed significant performance degradation during open-loop (\textbf{\texttt{OL}}) evaluations. 

 \textit{How can we adapt learned driving policies to distribution shifts using only a small amount of out-of-distribution~}(\textbf{\texttt{OOD}}) \textit{data?} Our analysis indicates that existing approaches suffer from catastrophic forgetting, leading to a degradation in performance when adapting to a new domain. To address this challenge, we introduce architectural and training procedure improvements specifically designed for structured policies.

The first generation of learning-based self-driving approaches used \emph{unstructured policies}, that directly approximate the entire decision-making process using a neural network, mapping a sequence of observations to an action~\cite{bojarski2016end, Kendall2019, Igl2022}. More recently, \emph{structured policies} have emerged, that maintain a modular yet differentiable architecture \cite{MP32021, UniAD2023, QuAD2024, DiffStack2022}, that offer various advantages such as verifiability, interpretability and better generalization while addressing challenges like compounding errors and information bottlenecks \cite{DiffStack2022}. For instance, planning \cite{Zeng_2019_CVPR, DiffStack2022, diehl2023corl, diehl2022differentiable, HuangDIPP2023} often employs differentiable optimization layers \cite{Amos2017,Amos2018} with learned cost functions. 
 
One of our key insights is that structured policies can also be better adapted to distribution shifts by adding residual layers to specific components. Specifically, we propose to add a \textit{Low-Rank Residual Decoder} (\textbf{\texttt{LoRD}}) in the structured policy during fine-tuning. For example, to adapt to different traffic norms in a new country, we can add a LoRD to the cost estimation network. Further, to mitigate \textit{catastrophic forgetting} \cite{MCCLOSKEY1989109} when fine-tuning on a small OOD dataset, we propose multi-task fine-tuning, where we train on a combination of in-distribution (\textbf{\texttt{ID}}) and OOD data.  

\begin{figure*}[ht]
    \centering
    \includegraphics[width=\linewidth]{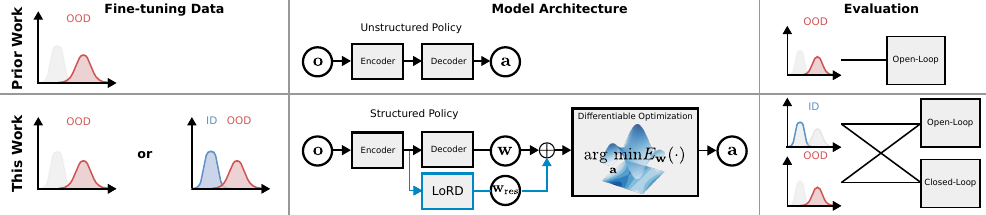}
    \vspace{-0.4cm}
    \caption{Overview of this work's contributions. Fine-tuning data: Mult-task fine-tuning with OOD and ID data. Architecture: LoRD predicts action or cost residuals. Evaluation: Open-loop and closed-loop evaluation in both domains (ID, OOD). $\mathbf{a}$: action, $\mathbf{o}$: sequence of observations, $\mathbf{E}_\mathbf{w}$: cost function, $\mathbf{w}$: cost parameters. }
    \label{fig::overview_contributions}
    \vspace{-0.4cm}
\end{figure*}

Previous works have proposed adaptation strategies such as fine-tuning with a few samples from the OOD domain (e.g., \cite{IvanovicICRA2023, MOSA}). However, most of these efforts focus on adapting individual tasks like trajectory prediction (unstructured policies) while overlooking integrating these tasks into a more complex, differentiable stack (structured policies) that includes prediction, planning, and control modules for SDVs (see Fig.~\ref{fig::overview_contributions}). Furthermore, there is a lack of research investigating the impact of these shifts on closed-loop (\textbf{\texttt{CL}}) performance, where the policy is executed rather than merely evaluated in an open-loop setting. Moreover, evaluations are often conducted exclusively in the OOD domain. We argue that evaluations should also be performed in the source (ID) domain to investigate forgetting and avoid the need for training a separate model for each domain. 

We evaluate our proposed methods on two real-world autonomous driving datasets (nuPlan, exiD), comparing them to prior work in both open-loop and closed-loop with cross-domain evaluations (ID and OOD). Our analysis shows that our method improves open-loop OOD while reducing forgetting by 23.33\% (ID) in comparison to fine-tuning (metric: bminSFDE) and improves the closed-loop OOD performance by 9.93 \% (metric: CL-NR).

Our specific contributions are as follows. First, we introduce a low-rank residual decoder that predicts residuals (e.g., cost parameters) for structured policies to adapt to the OOD domain. Second, we propose fine-tuning using a mix of in-distribution and OOD data to mitigate catastrophic forgetting. Third, we propose a comprehensive evaluation framework for SDV policy adaptation methods, including closed-loop metrics and both ID and OOD data. 

\section{Related Work}
\label{sec::related_work}

\textbf{SDV Policy Architectures.} 
Unstructured policies map from low-level sensor data \cite{Kendall2019, Jeffrey2020} or mid-level representations \cite{Igl2022} to actions. These approaches benefit from the absence of information bottlenecks and scalability with data, but their downside is the lack of interpretability \cite{DiffStack2022}. Trajectory prediction methods can also be viewed as unstructured policies as networks are trained via imitation learning \cite{MultiPath}.

Another research direction models policies in a more structured manner using differentiable algorithm networks \cite{Karkus2019}. These structured policies are also end-to-end trainable, while incorporating differentiable model-based components for perception \cite{Karkus2021}, prediction \cite{diehl2023icml, diehl2023corl, Liui2024}, and planning and control \cite{Zeng_2019_CVPR, DiffStack2022, diehl2022differentiable, HuangDIPP2023, QuAD2024}. However, no prior work has investigated the adaptation of structured policies to distribution shifts.

\textbf{Adaptation to Distribution Shifts.}
Distribution shifts can degrade model performance \cite{UniTraj}, leading to catastrophic failures. To mitigate this, \cite{Diehl2023RAL, diehl2021umbrella} penalize predicted OOD states using uncertainty estimates by ensembling learned models. The authors of \cite{Hallgarten2024, LLMAssist} propose using large language models to adapt parameters of non-differentiable planners at test time. Another line of transfer learning research focuses on fine-tuning networks with a few samples from the OOD domain. 
Various works \cite{IvanovicICRA2023, MOSA, Liu2022, Itkina2023, Ullrich_2024, Patrikar2024} have investigated adaptation in the trajectory prediction setting (open-loop), which could be considered as adapting unstructured policies whereas we adapt prediction, planning, and control and also evaluate in closed-loop. While it has been shown in the natural language processing (\textbf{\texttt{NLP}}) domain \cite{karimi2021parameterefficient} that multi-task fine-tuning can improve performance, this has not yet been explored for structured SDV policies.

\textbf{Residual Learning.}
Residual learning has been widely applied in fields such as computer vision \cite{ResNets,PA} and NLP to enhance model performance and  adaptation to distribution shifts. For instance, LoRA \cite{lora2023} is a commonly used parameter-efficient fine-tuning strategy for large language models \cite{lora2023}. In robotics, \cite{Silver2018} employs reinforcement learning to learn residual actions on top of an initial controller. In the context of trajectory prediction for autonomous driving, various works predict residuals on top of knowledge-driven model outputs \cite{Bahari2021} or anchors \cite{MultiPath}. In contrast, we utilize residuals to adapt the parameters of structured policies, such as cost functions.
The closest work to ours is \cite{MOSA}, which uses low-rank residuals \cite{lora2023} in the encoding network but focuses on adapting single-agent trajectories. In contrast, we employ a low-rank residual decoder to adapt cost parameters for multi-agent trajectories or cost weights. Additionally, in our application using structured policies, more interactions must be modeled in the decoders due to the multi-agent setting, motivating our design choice. This is later (Sec. \ref {seq::evaluation}) underlined by our results of LoRD outperforming \cite{MOSA}. Furthermore, we evaluate performance in closed-loop settings and conduct cross-evaluation in both the ID and OOD domains, as shown in Table \ref{tab:comparison}. 

\begin{table}[!h]
	\vspace{0.0cm}
	\centering
	\caption{A comparison to related works. TP: Trajectory Prediction, P: Planning, C: Control.}
	\resizebox{\columnwidth}{!}{
	\begin{tabular}{l c c c c c c c c c c c c}
		\toprule

		& Task & Adaptation& Fine-tuning& Structured & Residual & Residual & Eval. & Eval.    \\
		Method &  & & Data & Policy & Location & Adaptation & Domain & Setting   \\
		\midrule
		\cite{Bahari2021,MultiPath} & TP & \xmark & \xmark & \xmark & Decoder & Traj. & ID & OL   \\
		\cite{IvanovicICRA2023} & TP & \checkmark & OOD & \xmark &  \xmark & \xmark & OOD & OL  \\
		\cite{Ullrich_2024} & TP & \checkmark & OOD & \xmark &  \xmark & \xmark & OOD $\&$ ID & OL  \\
        \cite{Hallgarten2024, LLMAssist}  & P & \checkmark & \xmark & \xmark & \xmark & \xmark & OOD $\&$ ID & CL  \\
		\cite{MOSA}  & TP &\checkmark & OOD & \xmark & Encoder & Encoding & OOD & OL  \\ 
		\midrule
		Ours & TP,P,C & \checkmark &OOD, Mix & \checkmark & Decoder & Cost, Traj. & OOD $\&$ ID & OL $\&$ CL
		\\
		\bottomrule	
	\end{tabular}
    }
	\label{tab:comparison}
	\vspace{-0.4cm}
\end{table}

\section{Problem Formulation}

\textbf{Policy Representations.} Define the policy $\pi : \mathbf{o} \mapsto \mathbf{a}$  that maps from a sequence with length $H\in\mathbb{R}^{+}$ of observations $\mathbf{o} \in \mathbb{R}^{H \times n_o}$ to an action\footnote{We can extend this formulation to a multi-agent policy \cite{diehl2023corl}, whose notation is omitted for clarity.} $\mathbf{a}$. 
In this work, $\mathbf{a}$ is a control or state trajectory over a future horizon $T\in\mathbb{R}^{+}$, with $\mathbf{a} \in \mathbb{R}^{T \times n_a}$. The state and observation dimensions are denoted by $n_a$ and $n_o$, respectively. 
We can either choose an \textit{unstructured policy} $\mathbf{a} = \pi_\mathbf{w}(\mathbf{o})$ (explicit function) or a \textit{structured policy} \cite{florence22corl} (implicit function) defined by: 
$
\mathbf{a}^*=\arg \min _{\mathbf{a}} E_{\mathbf{w}}(\mathbf{a}, \mathbf{o})$, where the subscript $\textbf{w}$ denotes learnable parameters and $E_{\mathbf{w}}: \mathbb{R}^{T \times n_a} \times \mathbb{R}^{H \times n_o} \rightarrow \mathbb{R}$ is a cost function. We choose a neural network to represent the policy. We can also generalize this to a broader class $\mathbf{a}^*=\mathrm{A}(f_\mathbf{w}(o))$ where $A$ is an algorithm applied to the output of  a neural network $f$ with parameters $\mathbf{w}$.

Assume the neural network consists of an encoder-decoder structure as shown in Fig~\ref{fig::overview_contributions}. The encoder outputs a latent representation $\mathbf{z}$, which is then processed by the decoder to produce $\mathbf{y} = f_\mathbf{\theta}(\mathbf{z})$, where $\mathbf{\theta}$ represents the learnable parameters. The decoder may also include multiple heads that predict different outputs. In this work, $\mathbf{y}$ represents either an action $\mathbf{a}$ (for the unstructured policy) or parameters $\mathbf{w}$ (for constructing the cost $E_\mathbf{w}$) in the structured policy.
 
\textbf{Adapting to Distribution Shift.} We presume the availability of a \textit{base} model that has been pre-trained on the source ID domain. To account for a distribution shift and adapt this model to a new domain, we fine-tune the base model using a few samples from the target OOD domain. In principle, a shift between ID and OOD can occur, among others, due to different vehicle types, geographical locations, or weather conditions. 
 
For fine-tuning, we assume access to expert demonstrations in the form of a dataset  $D_E = \{\tau_k\}_{k=1}^{K_E}$ of $K_E\in \mathbb{N}^{+}$ trajectories $\tau_k = \{(\mathbf{o}_{k,1}, \mathbf{a}_{k,1}), (\mathbf{o}_{k,2}, \mathbf{a}_{k,2}), \dots, (\mathbf{o}_{k,T}, \mathbf{a}_{k,T})\}$ generated by an \textit{expert} policy $\pi_E$. We learn parameters by
\begin{equation}
    \mathbf{w}^* = \arg\min_{\mathbf{w}} \mathbb{E}_{(\mathbf{a}, \mathbf{o}) \sim p_{\mathrm{data}}}\left[L(\mathbf{w}; \mathbf{a}, \mathbf{o})\right],
\end{equation}
where \(L\) is a loss function, and \(p_{\mathrm{data}} = p_{\mathrm{OOD}}\) represents the data distribution.

\section{Methods}
\label{sec::methods}
We now describe our contributions in terms of architecture (residual decoder) and data (multi-task fine-tuning).

\subsection{Low-Rank Residual Decoder (LoRD)} 
\label{sec::low_rank_residual_decoder}

The main idea of our approach is that driving between different domains has similarities in various aspects (e.g., in terms of vehicle kinematics and agent behavior) but differs by a few features (e.g., left- vs. right-handed driving) \cite{MOSA}. For instance, most drivers try to avoid collisions while reaching a goal in a comfortable manner. We model this slight distribution shift using a \textit{residual decoder} $f_{\mathrm{res},\mathbf{\psi}}$ with parameters $\mathbf{\psi}$ that outputs a \textit{residual value} $\mathbf{y}_{\mathrm{res}} = f_{\mathrm{res},\mathbf{\psi}}(\mathbf{z})$, which is added to the output of the base decoder during adaptation:
\begin{equation}
    \hat{\mathbf{y}} = \mathbf{y} + \mathbf{y}_{\mathrm{res}}.
\end{equation}
Another hypothesized advantage of the residual is that the base network can maintain performance on the ID data, whereas adapting the weights of the residual decoder $f_{\mathrm{res},\mathbf{\psi}}$ can account for the distribution shift. Lastly, residuals provide a direct path for gradient flow, which helps mitigate the vanishing gradient problem during fine-tuning \cite{ResNets}.

In the case of a structured policy, the residual decoder of this work outputs \textit{residual cost parameters}\footnote{Also other outputs like system dynamic parameters or constraints used in a differentiable optimization \cite{Amos2018} could be predicted.} $\mathbf{w}_{\mathrm{res}}$ to parameterize a \textit{residual cost} $E_{\mathrm{res}}$. For an unstructured policy, the residual decoder outputs a \textit{residual action} $\mathbf{a}_{\mathrm{res}}$.

While $f_{\mathrm{res},\mathbf{\psi}}$ can have various structures, this work uses a single linear layer without bias to minimize the number of additional parameters. Inspired by the fine-tuning of large language models \cite{lora2023} and applications in marginal trajectory prediction \cite{MOSA}, we perform a low-rank matrix decomposition 
\begin{equation}  
 f_{\mathrm{res},\psi} = \mathbf{B}\mathbf{A},  
 \label{equ::low_rank_decomposition}
\end{equation}
leading to a further parameter reduction and computational efficiency. 
$\mathbf{B} \in \mathbb{R}^{n_B \times r}$ and $\mathbf{A} \in \mathbb{R}^{r \times n_A }$ describe matrices with rank $r \ll \min(n_A, n_B)$. The matrices are initialized such that the residuals initially do not influence the original network, which stabilizes the training \cite{lora2023}. We also perform dropout on the input of $f_{\mathrm{res},\mathbf{\psi}}$ with probability $p_\mathrm{drop}$ for better generalization. This decoder is referred to as \textit{Low-Rank Residual Decoder}.

Although residuals can also be added in the encoder \cite{MOSA,PA}, we opt to add them in the decoder because differentiable joint prediction and planning methods \cite{diehl2023corl,HuangDTPP2024} model interactions, occurring in multi-agent driving scenarios, explicitly in the decoders and differentiable optimization layers. 

\subsection{An Energy-based Model Perspective} 
Next, we highlight a connection of the proposed LoRD to the compositional properties \cite{YilunDu2020,YilunDu2021} of energy-based models (EBM) \cite{lecun06} in the case of structured policies. We show this perspective, as it allows for a principled way of integrating structured policies with residual adaptations.

We begin by interpreting the cost as an energy \cite{GANIRLEBM} and assuming a probabilistic policy $p_{\mathbf{w}}(\mathbf{a}|\mathbf{o})$. One possible way to turn energies into probabilities is by using the Gibbs distribution $ p_{\mathbf{w}}(\mathbf{a}|\mathbf{o}) = \frac{\exp\left(-E_\mathbf{w}(\mathbf{a},\mathbf{o})\right)}{Z_\mathbf{w}(\mathbf{o})} $, with an underlying partition function $ Z_\mathbf{w}(\mathbf{o}) = \int \exp\left(-E_\mathbf{w}(\mathbf{a},\mathbf{o})\right) \textrm{d}\mathbf{a}$. Other EBM learning techniques can be found in \cite{lecun06, diehl2023icml}. Prior work \cite{YilunDu2020,YilunDu2021,Comas2023} has shown how to compose energies using a product of experts \cite{hinton1999products} $\prod_{j} p_{\mathbf{w}}^{j}(\mathbf{a},\mathbf{o}) \propto e^{-\sum_{j} E_\mathbf{w}^{j}(\mathbf{a},\mathbf{o})}$.

In the above described setting, LoRD predicts $\mathbf{w}_{\mathrm{res}}$ used to parametrize a cost/energy $E_{\mathbf{w}_{\mathrm{res}}}$ resulting in 
$p_{\mathbf{w}, \mathbf{w}_\mathrm{res}}(\mathbf{a},\mathbf{o}) \propto \exp(-\left(E_{\mathbf{w}}(\mathbf{a},\mathbf{o}) + E_{\mathbf{w}_{\mathrm{res}}}(\mathbf{a},\mathbf{o})\right)$.

Hence, under our formulation in Sec. \ref{sec::low_rank_residual_decoder}, \textit{we aim to find actions $\mathbf{a}$ that have high likelihood under a composition of the old distribution (ID) and a residual distribution}, whereas the weighting of both is learned. For instance, a lane-keeping action that was highly probable on highways (ID) may also need to account for new residual factors such as narrower streets in an urban environment (residual distribution).

\subsection{Multi-Task Fine-Tuning}
Fine-tuning with data from the OOD target domain can lead to catastrophic forgetting of the original ID domain \cite{MCCLOSKEY1989109}. To mitigate this issue, we propose using multi-task fine-tuning, which facilitates information sharing across various tasks and has demonstrated effectiveness in the natural language processing domain \cite{karimi2021parameterefficient}. Specifically, during fine-tuning, we sample data from a distribution $ p_{\mathrm{MT}} = p_{\mathrm{ID}} + \alpha p_{\mathrm{OOD}} $, where $\alpha$ represents the ratio of ID to OOD data.

\section{Empirical Evaluation}
\label{seq::evaluation}

This section investigates the following research questions: \textit{Q1}:  How do LoRD and multi-task fine-tuning \textbf{\texttt(FT)} impact the open-loop and closed-loop performance in ID and OOD scenarios compared to the state-of-the-art baselines?\\
\textit{Q2}: Are structured policies (\textbf{\texttt{SP}}) and unstructured policies (\textbf{\texttt{USP}}) impacted by adding LoRD?

\begin{table*}[!t]
	\centering
	\caption{Open-loop joint prediction results on exiD. We report the \textit{mean and standard deviation} of the metric over three runs, which were initialized with different random training seeds.} 
	\resizebox{\textwidth}{!}{  
		\begin{tabular}{l | c | c | c | c c c c | c c c c}
			\toprule 
			& & & & \multicolumn{4}{c}{OOD} & \multicolumn{4}{c}{\IDOOD }\\
			\cmidrule(r){5-8}
			\cmidrule(r){9-12}
			Method & Residual & FT Layer & FT Data & $\mathrm{minSADE}\downarrow$ & $\mathrm{minSFDE}\downarrow$ & $\mathrm{bminSFDE}\downarrow$ & $\mathrm{SMR}\downarrow$ & $\mathrm{minSADE}\downarrow$ & $\mathrm{minSFDE}\downarrow$ & $\mathrm{bminSFDE}\downarrow$ & $\mathrm{SMR}\downarrow$ \\
			\midrule

			Base \cite{diehl2023corl} & \xmark & \xmark & \xmark & $1.05\pm0.23$ & $2.55\pm0.05$  & $3.17\pm0.05$ & $0.15\pm0.01$ & $0.91\pm0.12$ & $2.25\pm0.05$  & $2.84\pm0.02$ & $0.11\pm0.01$\\ 
			\cmidrule(r){1-12} 
			\footnotesize MOSA (A+F) \cite{MOSA} & Encoder & Residual & OOD & $0.96\pm0.02$ & $2.29\pm0.06$  & $2.93\pm0.07$ & $0.13\pm0.02$ & $1.00\pm0.03$  & $2.50\pm0.10$ & $3.13\pm0.04$ & $0.16\pm0.03$\\ 
			MOSA (F) \cite{MOSA}  & Encoder & Residual & OOD & $0.95\pm0.01$ & $2.27\pm0.01$  & $2.90\pm0.02$ & $0.11\pm0.00$ & $\boldsymbol{0.95\pm0.00}$  & $2.37\pm0.01$ & $3.00\pm0.03$ & $0.13\pm0.01$\\   
			
			Parallel Adapter  \cite{PA}  & Encoder & All & OOD & $0.89\pm0.00$ & $2.16\pm0.01$  & $2.82\pm0.05$ & $0.09\pm0.01$ & $1.01\pm0.04$ & $2.53\pm0.12$ & $3.21\pm0.04$ & $0.15\pm0.00$ \\
			Parallel Adapter  + LoRD (Ours)  & Both & All & OOD & $\boldsymbol{0.87\pm0.01}$ & $\boldsymbol{2.07\pm0.03}$  & $2.71\pm0.01$ & $\boldsymbol{0.07\pm0.01}$ & $1.00\pm0.01$ & $2.50\pm0.02$ & $3.17\pm0.02$ & $0.14\pm0.01$ \\  
			
			Partial Fine-tuning   & \xmark & Last & OOD & $0.88\pm0.01$ & $2.16\pm0.02$  & $2.82\pm0.05$ & $0.09\pm0.00$ & $1.00\pm0.02$ & $2.50\pm0.04$ & $3.13\pm0.04$ & $0.15\pm0.01$ \\
			Fine-tuning   & \xmark & All  & OOD  &$0.88\pm0.01$ & $2.12\pm0.02$  & $2.77\pm0.01$ & $0.09\pm0.01$ & $1.00\pm0.02$ & $2.51\pm0.05$ & $3.17\pm0.02$ & $0.15\pm0.02$\\ 
			Fine-tuning + LoRD (Ours) & Decoder & All & OOD & $0.88\pm0.02$ & $2.11\pm0.04$  & $\boldsymbol{2.71\pm0.03}$ & $0.09\pm0.00$ & $0.98\pm0.01$ & $\boldsymbol{2.36\pm0.03}$ & $\boldsymbol{2.95\pm0.09}$ & $\boldsymbol{0.12\pm0.00}$ \\
            \midrule
			\midrule
			
			Partial Fine-tuning      & \xmark & Last  & Mix& $0.91\pm0.01$ & $2.21\pm0.01$  & $2.85\pm0.01$ & $0.10\pm0.00$ & $0.85\pm0.01$ & $2.13\pm0.01$ & $2.76\pm0.01$ & $\boldsymbol{0.08\pm0.01}$ \\ 
			Fine-tuning        & \xmark & All & Mix & $0.86\pm0.01$ & $2.08\pm0.05$  & $2.68\pm0.09$ & $0.09\pm0.01$ & $0.86\pm0.01$ & $2.15\pm0.04$ & $2.76\pm0.06$ & $0.10\pm0.01$\\ %
			Fine-tuning + LoRD (Ours) & Decoder & All & Mix & $\boldsymbol{0.86\pm0.01}$ & $\boldsymbol{2.07\pm0.03}$  & $\boldsymbol{2.68\pm0.05}$ & $\boldsymbol{0.09\pm0.00}$ & $\boldsymbol{0.84\pm0.01}$ & $\boldsymbol{2.01\pm0.02}$ & $\boldsymbol{2.72\pm0.05}$ & $0.09\pm0.00$ \\

			\bottomrule
		\end{tabular}

	}
	\raggedright
\label{tab::main_results_exiD}
\end{table*}

\begin{table*}[!t]
    \vspace{-0.1cm}
    \centering
    \caption{Open-loop planning (left) and closed-loop control (right) results on nuPlan test scenarios. Closed-loop regularization denotes training with reward regularization and history dropout.}

    \resizebox{\textwidth}{!}{%
        \begin{tabular}{l|c|c|c|c|cccc||cccc}
            \toprule
            &&&&& \multicolumn{4}{c}{Open-Loop} & \multicolumn{4}{c}{Closed-Loop} \\
            &&&&Closed-Loop& \multicolumn{2}{c}{OOD} & \multicolumn{2}{c}{\IDOOD } &\multicolumn{2}{c}{OOD} & \multicolumn{2}{c}{\IDOOD } \\
            \cmidrule(r){6-7}
            \cmidrule(r){8-9}
            \cmidrule(r){10-11}
            \cmidrule(r){12-13}
            Method & Residual & FT Layer & FT Data & Regularization &
            $\mathrm{ADE}\downarrow$ & $\mathrm{FDE}\downarrow$ & $\mathrm{ADE}\downarrow$ & $\mathrm{FDE}\downarrow$ &
            $\mathrm{CL}$-$\mathrm{NR}\uparrow$ & $\mathrm{CL}$-$\mathrm{R}\uparrow$ & $\mathrm{CL}$-$\mathrm{NR}\uparrow$ & $\mathrm{CL}$-$\mathrm{R}\uparrow$ \\
            \midrule
            Base \cite{HuangDTPP2024} (SP) & \xmark & \xmark & \xmark & \xmark & $3.32$ & $6.73$ & $3.05$ &$5.95$ & $0.740$ & $0.705$ & $0.758$ & $0.718$ \\
            \cmidrule(r){1-13}
            
            Fine-tuning &\xmark &All & OOD& \xmark  & $2.53$ & $\boldsymbol{5.04}$ & $2.75$ &$5.65$ &$0.667$ & $0.695$ & $0.611$& $0.669$ \\
            Fine-tuning + LoRD (Ours)&Decoder&All & OOD & \xmark & $2.42$ & $5.05$ & $2.64$ &$5.50$& $0.655$ & $0.678$ & $0.606$& $0.683$ \\
            Fine-tuning + LoRD (Ours)&Decoder&All & Mix & \xmark  & $\boldsymbol{2.37}$ & $5.58$ & $\boldsymbol{2.47}$ &$\boldsymbol{5.47}$ &$0.708$ & $0.695$ & $0.713$& $0.655$ \\
            Fine-tuning &\xmark &All& OOD & \checkmark & $2.54 $ & $5.34$ & $2.71$& $5.80$ &$0.703 $ & $0.702$ & $0.702 $& $0.644$ \\
            Fine-tuning + LoRD (Ours)&Decoder&All & OOD& \checkmark &$2.54$ & $5.34$ & $2.70$ & $5.82$ &$\mathbf{0.759}$ & $\mathbf{0.764}$ & $\mathbf{0.728}$ & $\mathbf{0.695}$ \\
            \midrule
            \midrule
            Base (USP) &\xmark &\xmark & \xmark & \xmark & $3.34$ & $4.53$ & $2.61$ &$5.06$ & $0.105$ & $0.115$ & $0.123$ & $0.150$ \\
            \cmidrule(r){1-13}
            Fine-tuning & \xmark &All & OOD&\xmark  & $\boldsymbol{1.47}$ & $\boldsymbol{4.41}$ & $\boldsymbol{1.95}$ &$5.96$ &$\boldsymbol{0.194}$ & $0.178$ & $\boldsymbol{0.176}$& $0.167$ \\
            Fine-tuning + LoRD (Ours) & Decoder &All& OOD& \xmark &$1.52$ & $4.47$ & $1.96$ &$\boldsymbol{5.87}$ &$0.174$ & $\boldsymbol{0.214}$ & $0.155$& $\boldsymbol{0.209}$ \\
			\bottomrule
        \end{tabular}
    }
    \label{tab:nuPlan_openclosedloop}
    \vspace{-0.3cm}
\end{table*}

Our key findings are the following: \textbf{F1}: LoRD performs similar or better than the baselines (OOD) and keeps the ID performance better by a large margin in open-loop. Multi-task fine-tuning improves the performance of all methods. \textbf{F2}: While the trend for SP is the same as in open-loop and LoRD is the best adaptation method, we observe a large gap between open-loop and closed-loop performance, which is in line with prior work \cite{DaunerCoRL2023, zhai2023}. Moreover, the fine-tuning data distribution has a high impact in closed-loop. \textbf{F3}: Unstructured policies benefit less using LoRD. \textbf{F4}: All design choices contribute to the performance.

\subsection{Experimental Setup}
\label{sec::experimental_setup}
\textbf{Datasets and Evaluation Environments}.
\textbf{\texttt{exiD}} is an interactive highway dataset captured in Germany that contains recordings of seven locations with no geographic overlap. We split the data such that the ID splits contain data from the same six maps but from different sequences. Recordings of the last map are preserved for the OOD split. One sample constitutes of $\SI{4}{\second}$ future with $\SI{2}{\second}$ of history sampled at $\Delta t=\SI{0.2}{\second}$. The ID sets contain 110431 (train) / 12425 (val) / 13884 (test) samples from different sequences. The fine-tuning OOD sets contain 12378 (train) / 3459 (val) / 3561 (test) samples. 
\textbf{\texttt{nuPlan}} \cite{nuPlan} is a standard benchmark for motion planning in urban driving environments, enabling learning using large-scale data from different geographies. We use data from Boston and Pittsburgh (ID, right-handed traffic) to train a \textit{base} model and data from Singapore (OOD, left-handed traffic) for fine-tuning. We evaluate models in both domains (ID, OOD), in open-loop and closed-loop. During closed-loop evaluation, the planned trajectory is executed. Data is sampled with $\Delta t=\SI{0.1}{\second}$, and the future trajectory and agent history length are $\SI{8}{\second}$ and $\SI{2}{\second}$, respectively. The ID sets contain 155366 (train) / 21853 (val) / 22449 (test) samples split by sequences. The fine-tuning OOD set contains 32,901 (train) / 7067 (val) / 9182 (test) samples. We extract 80 unseen scenarios for closed-loop evaluation per domain (ID or OOD) accounting for balancing scenario types.

While our approach can potentially be used for any kind of distribution shift, we test it for geographical shifts as these are easy to simulate with real-world data.

\textbf{Models and Policies}. 
We evaluate our contributions on top of two different base models: On exiD, the model consists of \textbf{\texttt{EPO}} \cite{diehl2023corl}, a differentiable game-theoretic gradient-based optimization approach for joint multi-agent predictions and control (structured multi-agent policy). We apply the residual on all predicted game parameters (cost weights and agents' goals) and the initial joint control trajectories.
The nuPlan experiments use \textbf{\texttt{DTPP}} \cite{HuangDTPP2024}, a structured policy. During learning the approach differentiates through sampling-based optimization. We use the integration in \texttt{tu\_garage} \cite{DaunerCoRL2023} for closed-loop simulation provided by \cite{Hallgarten2024}. The residual is added to the predicted agents' trajectories used in the SDV cost function. We further explore extensions of DTPP, training with additional rewards, and history dropout (see \textbf{F2}), which are specifically useful for closed-loop control.

\textbf{Baselines}. 
We compare to the following baselines: 
\begin{itemize}
\item \textit{FT}: Fine-tuning of all parameters as in \cite{IvanovicICRA2023}. 
\item \textit{PFT}: Partial model fine-tuning adapting the last layer of each decoder as in \cite{IvanovicICRA2023}. 
\item \textit{MOSA} \cite{MOSA}: MOSA employs low-rank adaptation layers in the encoder while freezing the base networks. MOSA (F) applies adapters in the attention-based fusion layers from \cite{diehl2023corl}, and MOSA (A+F) adds adapters in the agent encoders. 
\item \textit{PA} \cite{PA}: Parallel adapter similar to MOSA but without freezing of weights. 
\end{itemize}
On exiD (prediction) we compare extensively against prior methods, while for the planning and control experiments on nuPlan we focus on the open-loop to closed-loop gap and compare against the strongest baseline.
 
\textbf{Hyperparameters}. 
\textit{To ensure a fair comparison, we performed grid searches for the hyperparameters (learning rate, dropout probability, rank in Equ. (\ref{equ::low_rank_decomposition})) of all our models, baselines, and ablations}. 
For open-loop evaluation we use the checkpoint with the lowest validation metrics (exiD: minSADE, nuPlan: ADE). For closed-loop the last checkpoint is chosen as our preliminary experiments showed that this correlated better with closed-loop performance, which is in line with related work \cite{Hussenot2021}.

\textbf{Metrics}. For closed-loop evaluation, we use the official CL-scores (1: best score, 0: worst score) of the nuPlan simulation, where \textbf{\texttt{CL-R}} denotes a reactive and \textbf{\texttt{CL-NR}} a non-reactive (log-replay) simulation of the other agents. The open-loop evaluation uses the following standard metrics. 
On nuPlan, we use the \textit{Average Displacement Error} (\textbf{\texttt{ADE}}) and \textit{Final Displacement Error} (\textbf{\texttt{FDE}}) of the SDV's trajectory. For joint predictions (exiD), the \textbf{\texttt{minSADE}} and \textbf{\texttt{minSFDE}} are the multi-modal scene-level equivalents \cite{diehl2023corl}. The \textbf{\texttt{bminSFDE}} (official ranking metric of Argoverse 2 \cite{Argoverse2}) penalizes joint predictions with poorly calibrated probabilities. Lastly, we compute the \textit{Scene Miss Rate} (\textbf{\texttt{SMR}}), which describes the percentage of samples where minSFDE $ \geq \SI{3.6}{\meter}$. We report the performance in the OOD domain and the average performance over the ID and OOD domain denoted by \IDOOD \footnote{A model that does not adapt during fine-tuning (bad OOD performance) would still have good ID performance, motivating to report \IDOOD. ID performance can be computed by $\text{ID} = 2* \IDOOD - \text{OOD}$}. 

\subsection{Results}
\label{sec::results}
\begin{figure*}[ht]
    \centering
    \begin{subfigure}{0.20\textwidth}
      \includegraphics[width=\linewidth]{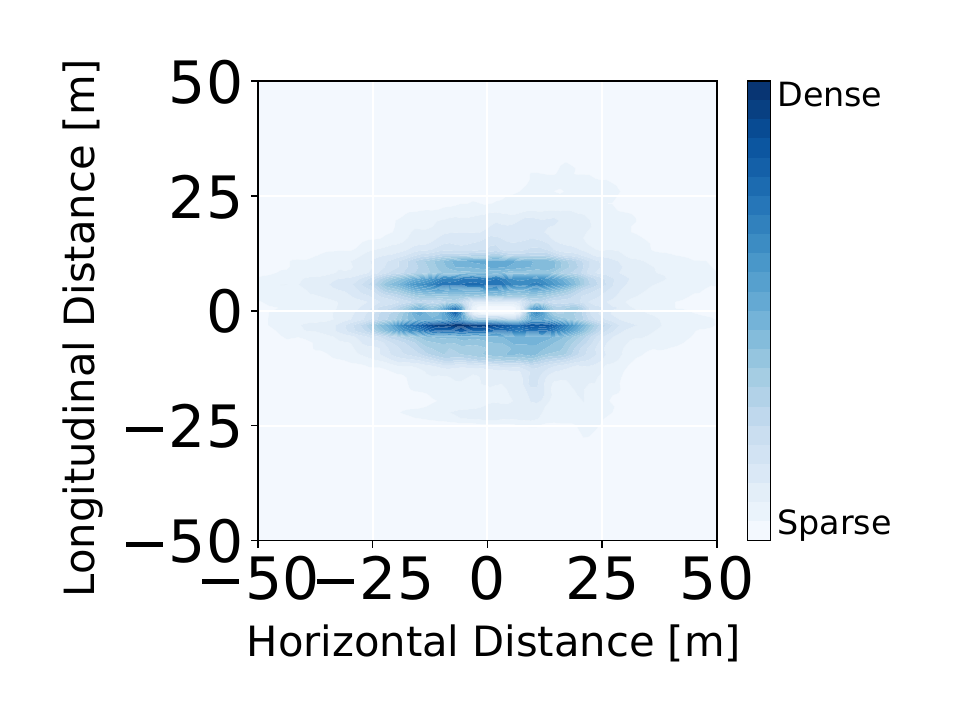}
      \vspace{-0.6cm}
      \caption{}
      \label{fig:subfig1}
    \end{subfigure}
    \begin{subfigure}{0.20\textwidth}
      \includegraphics[width=\linewidth]{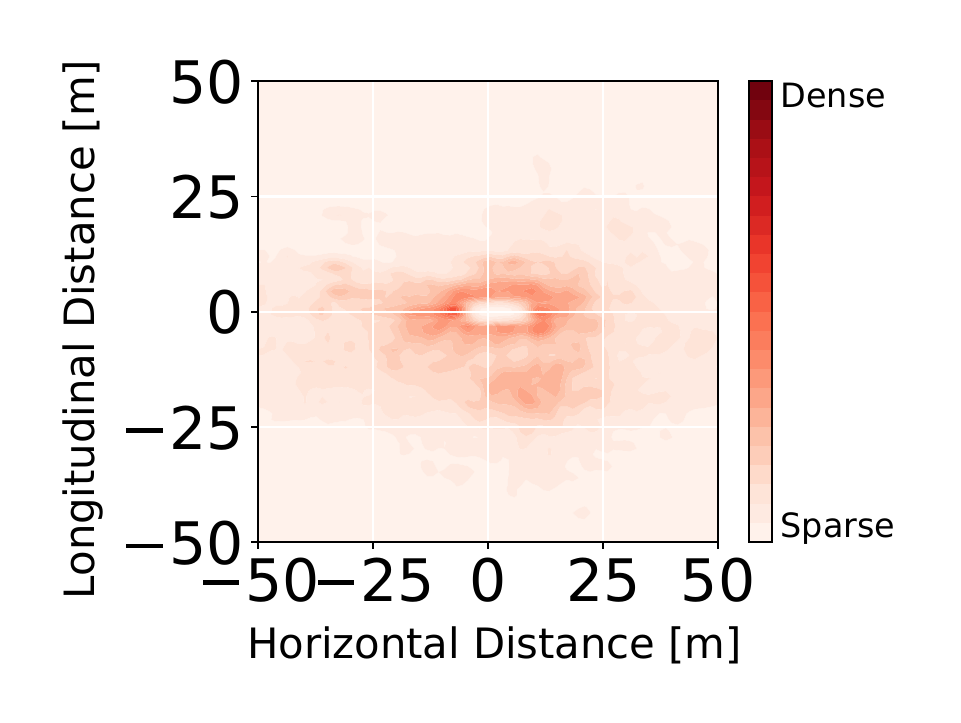}
      \vspace{-0.6cm}
      \caption{}
      \label{fig:subfig2}
    \end{subfigure}
    \begin{subfigure}{0.19\textwidth}
      \includegraphics[width=\linewidth]{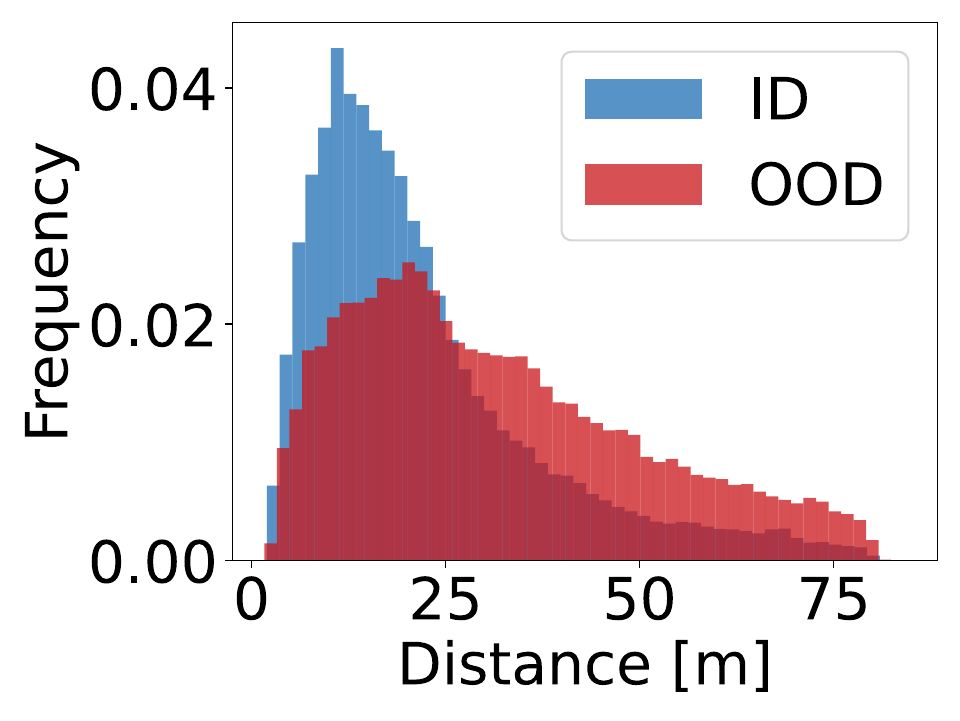}
      \vspace{-0.6cm}
      \caption{}
      \label{fig:subfig3}
    \end{subfigure}
    \begin{subfigure}{0.19\textwidth}
      \includegraphics[width=\linewidth]{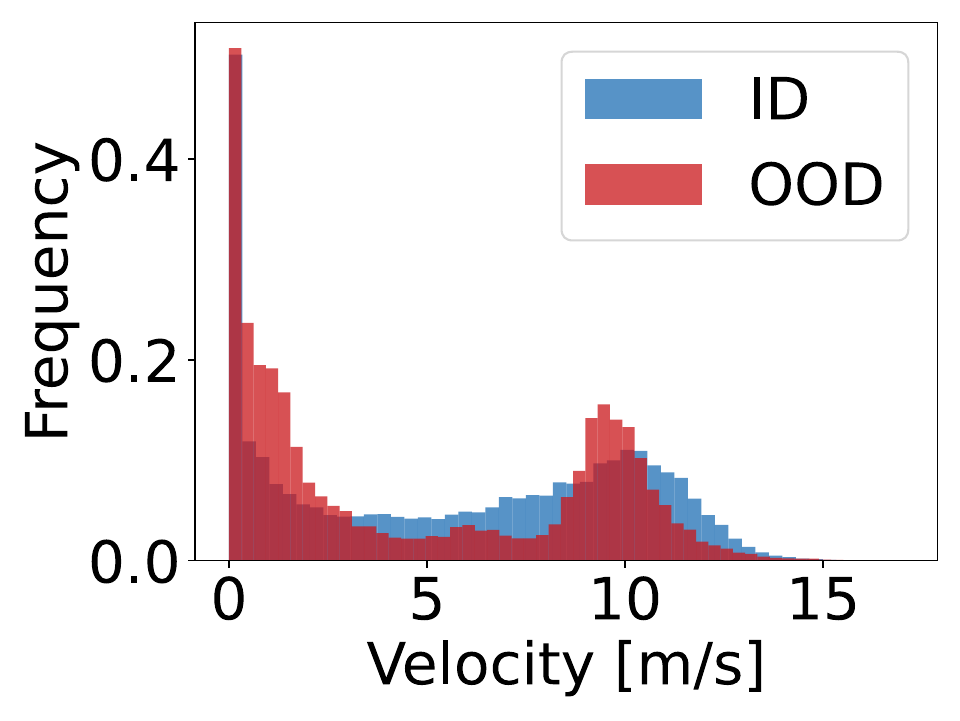}
      \vspace{-0.6cm}
      \caption{}
      \label{fig:subfig4}
    \end{subfigure}
    \begin{subfigure}{0.19\textwidth}
        \includegraphics[width=\linewidth]{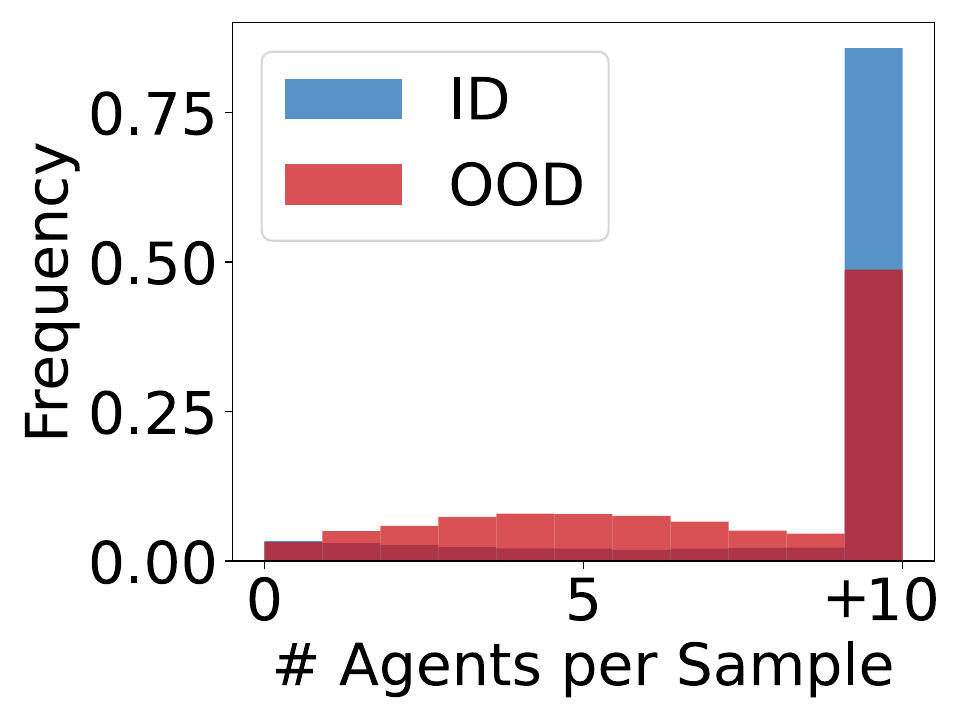}
        \vspace{-0.6cm}
        \caption{}
    \label{fig:}
    \end{subfigure}  
    \caption{Dataset statistics of the nuPlan ID (Boston, Pittsburgh) and OOD (Singapore) domain. Spatial distribution of distances to other agents using kernel density estimates (a,b). Distributions of the Euclidean distances to other agents (c), velocities (d), and number of agents (e). }
    \label{fig::dataset_analysis}
  \vspace{-0.4cm}
  \end{figure*}

\begin{figure}[htbp]
    \centering
    \vspace{-0.0cm}
    \includegraphics[trim={2.5cm 5.5cm 2.5cm 2.5cm},clip, width=0.45\linewidth]{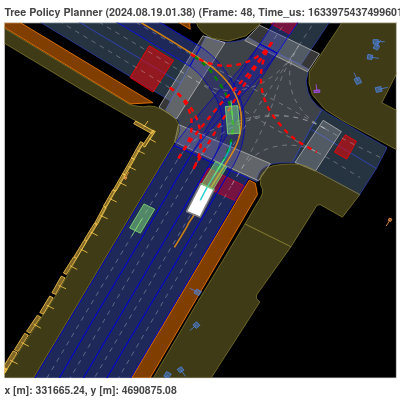} 
    \includegraphics[trim={2.5cm 5.5cm 2.5cm 2.5cm},clip, width=0.45\linewidth]{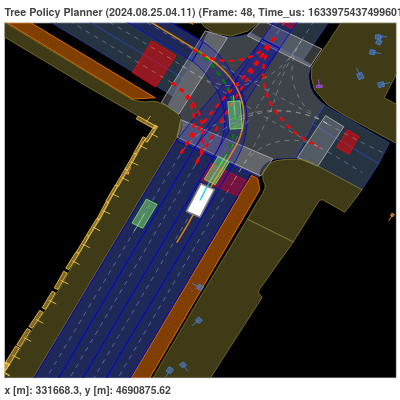} 

    \caption{Qualitative results (closed-loop control). The SDV (white) using the FT baseline (left) causes a collisions with another vehicle (green). The FT + LoRD (closed-loop regularization, right) model brakes and avoids the collision evident by the length of the cyan trajectory. }
    \label{fig:qualitative_closed_loop_comparison}
    \vspace{-0.4cm}
\end{figure}

\begin{figure}[] 
	\vspace{-0.0cm}
	\centering
\begin{tikzpicture}

\definecolor{darkgray176}{RGB}{176,176,176}
\definecolor{lightgray204}{RGB}{204,204,204}


\definecolor{reds_middle}{RGB}{216,81,85}
\definecolor{blues_middle}{RGB}{88,148,199}

\begin{axis}[
legend cell align={left},
legend style={
  fill opacity=0.8,
  draw opacity=1,
  text opacity=1,
  at={(0.91,0.5)},
  anchor=east,
  draw=lightgray204
},
width=0.9\columnwidth,
height=1.28in,
tick align=outside,
tick pos=left,
x grid style={darkgray176},
xlabel={Added ID data [\(\displaystyle \%\)]},
xmajorgrids,
xmin=-5, xmax=105,
xtick style={color=black},
y grid style={darkgray176},
ylabel={ADE\(\displaystyle _{\mathrm{plan}}\) [m]},
ymajorgrids,
ymin=2.34955, ymax=2.88745,
ytick style={color=black}
]
\addplot [thick, blues_middle, mark=*, mark size=4, mark options={solid}]
table {%
0 2.863
25 2.557
50 2.768
100 2.79
};
\addlegendentry{ID}
\addplot [thick, reds_middle, mark=*, mark size=4, mark options={solid}]
table {%
0 2.423
25 2.374
50 2.444
100 2.47
};
\addlegendentry{OOD}
\end{axis}

\end{tikzpicture}
	\vspace{-0.3cm}
	\caption{Ablation study on the amount of added ID data during multi-task fine-tuning. 0\% denotes standard fine-tuning. 100\% adds the same amount of ID and OOD data.}
	\label{fig::ablationsteps}	
	\vspace{-0.5cm}
\end{figure}
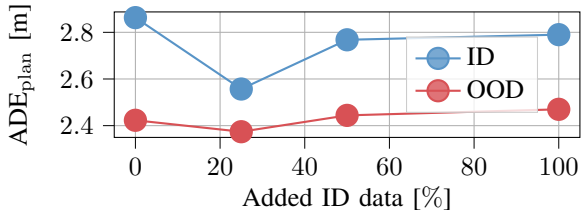

Our findings, are visualised in Tab. \ref{tab::main_results_exiD} (exiD, open-loop) and \ref{tab:nuPlan_openclosedloop} (nuPlan, open-loop and closed-loop). Fig. \ref{fig::dataset_analysis} and Fig. \ref{fig:qualitative_closed_loop_comparison} provide a dataset analysis and qualitative results for nuPlan.  Lastly,  Fig. \ref{fig::ablationsteps} and Tab. \ref{tab:ablations_results_exiD} present ablation studies.

\textbf{F1: Improvements due to LoRD and multi-task fine-tuning for structured policies.}
Tab. \ref{tab::main_results_exiD} and \ref{tab:nuPlan_openclosedloop} (upper-left) compare our methods against baselines considering open-loop joint prediction (exiD) or planning (nuPlan) metrics. We observe that the performance of all SP base models \cite{HuangDTPP2024,diehl2023corl} drop in the OOD scenarios, which is in line with the recent trajectory prediction study \cite{UniTraj}. After fine-tuning, the OOD performance of all methods improves, while their ID performance declines. However, methods incorporating LoRD are the least affected in ID performance. Adding LoRD to various methods results in better or similar performance in OOD scenarios and  outperforms the baselines in \IDOOD \, metrics.
MOSA, which is similar to LoRD but adds residuals in the encoding, performs worse. That can be explained by the fact that MOSA focuses on marginal trajectory prediction while freezing the base network. It was previously shown \cite{Makansi2022} that models are often unable to reason about interactions in the encoding and learn to extrapolate kinematics. In contrast, our work jointly predicts all agents' trajectories, where interactions are explicitly modeled in the decoders and differentiable optimization layers. 

\textit{By adding ID data during multi-task  fine-tuning, we observe that the effect of forgetting is reduced across all methods}, evident by the better performance in \IDOOD \,metrics. For instance, the minSFDE ID of the fine-tuning baseline is decreased by $\SI{0.67}{\metre}$ (23.33\%) in the exiD experiments using our approach combining LoRD with multi-task fine-tuning, which performs best on both datasets in open-loop.

\textbf{F2: Gap between open-loop and closed-loop performance.}
We consider the adaptation methods applied to the SP base model (DTPP \cite{HuangDTPP2024}). Interestingly, \textit{fine-tuning using OOD (Singapore) data decreases the OOD performance} (Tab. \ref{tab:nuPlan_openclosedloop}, upper-right) for both the baseline and our contribution. One hypothesis for this effect is that it occurs due to the characteristics of the Singapore data. To provide evidence for that claim, we perform a dataset analysis of the Boston and Pittsburgh (ID), as well as Singapore (OOD) training datasets visualized in Fig. \ref{fig::dataset_analysis}. We observe that Singapore contains more samples with higher distances to other agents (a-c). Moreover, velocities are lower (d), and samples contain fewer agents (e) in line with findings in \cite{Patrikar2024}. That indicates a reduced complexity in the Singapore dataset, which can elevate the causal confusion \cite{deHaanNeurIPS2019} effect. For instance, when a scene is mostly empty, future observations can be better explained by the SDV's history, and the model learns not to account for other traffic participants.

To account for this reduced complexity, one could up-sample specific difficult scenarios by training a difficulty model like \cite{Bronstein2023}. Alternatives to improve OOD performance are to use additional rewards \cite{Lu2023} or a history dropout \cite{chengplanTF2024}. We apply both latter techniques (denoted by closed-loop regularization in Tab. \ref{tab:nuPlan_openclosedloop}) and observe that the performance of the fine-tuned models increases. In particular, we added a progress and collision reward loss during training.
\textit{Adding the LoRD achieves the overall best performance of the fine-tuning methods, consistent with results in the open-loop evaluation}. Fig. \ref{fig:qualitative_closed_loop_comparison} visualizes a qualitative closed-loop comparison. The fine-tuning baseline underestimates the slow movement of the leading agent, evidenced by the length of the planned trajectory, and causes a rear collision. In contrast, our FT + LoRD model reacts accordingly with a breaking maneuver and avoids the collision. Further results across different traffic norms in the regions are in the supplementary video.

\textbf{F3: Unstructured policy does benefit less from LoRD.}
We also apply the LoRD to an an unstructured policy, and the residual is added to the SDV trajectory. The model uses the same encoder as \cite{HuangDTPP2024} but predicts unimodal trajectories for the SDV and other agents using two multi-layer perceptrons. That architecture is inspired by \cite{chengplanTF2024} but differs slightly from the original implementation. Moreover, we use only about 11\% of the data as \cite{chengplanTF2024}. Hence, the results are not directly comparable to those in the original work.

Consider the open-loop nuPlan experiments using the USP base model in Tab. \ref{tab:nuPlan_openclosedloop} (bottom-left). We make the following observations. First, the performance is similar, and the fine-tuning baseline even outperforms the model with added LoRD in three out of four metrics. Hence, we see no significant improvement due to adding LoRD.   

Second, in open-loop, the fine-tuned USP models outperform the SP in OOD scenarios. For instance, the difference in ADE is approximately $\SI{0.9}{\metre}$. This can be attributed to the USP being trained via regression. In contrast, the SP learns through classification and must select from a finite set of sampled trajectories, restricting the solution space.

 However, consider the difference in closed-loop performance in Tab. \ref{tab:nuPlan_openclosedloop} (left vs right). Comparing the USP and SP models, it becomes apparent that the observations are opposite to those in open-loop evaluation: the SP base model outperforms the USP base model by a large margin. This improvement is partly due to the additional structure introduced by sampling lane-aligned trajectories reducing the risk of off-road collisions. Moreover, this structure enhances sample efficiency, which is particularly beneficial in the data regime used in this work. Lastly, the USP is more sensitive to distribution shifts, which frequently occur due to compounding errors, due to model training via offline imitation learning \cite{Ross10}.  
 This again underlines the finding \textbf{F2}, showing the gap between OL and CL performance. 

\begin{table}[!ht]
    \vspace{-0.2cm}
    \centering
    \caption{Ablation study on exiD.}
    \resizebox{\columnwidth}{!}{%
        \begin{tabular}{lccccc}
            \toprule
            & \multicolumn{2}{c}{OOD} & \multicolumn{2}{c}{\IDOOD} \\
            \cmidrule(r){2-3} \cmidrule(r){4-5}
            Method & $\mathrm{minSFDE}\downarrow$ & $\mathrm{bminSFDE}\downarrow$ & $\mathrm{minSFDE}\downarrow$ & $\mathrm{bminSFDE}\downarrow$ \\
            \midrule
            Fine-tuning + LoRD (Base) & $\boldsymbol{2.10 \pm 0.05}$ & $\boldsymbol{2.70 \pm 0.03}$ & $\boldsymbol{2.34 \pm 0.02}$ & $\boldsymbol{2.96 \pm 0.03}$ \\
            No goal residual & $2.10 \pm 0.03$ & $2.71 \pm 0.05$ & $2.46 \pm 0.05$ & $3.09 \pm 0.06$ \\
            No cost residual & $2.10 \pm 0.01$ & $2.75 \pm 0.01$ & $2.44 \pm 0.06$ & $3.09 \pm 0.07$ \\
            No low-rank  & $2.12 \pm 0.01$ & $2.76 \pm 0.04$ & $2.44 \pm 0.04$ & $3.08 \pm 0.04$ \\
            Residual on output & $2.09 \pm 0.02$ & $2.72 \pm 0.01$ & $2.46 \pm 0.02$ & $3.10 \pm 0.07$ \\
            \bottomrule
        \end{tabular}
    }
    \label{tab:ablations_results_exiD}
    \vspace{-0.2cm}
\end{table}

\textbf{F4: Ablation Study.}
 We ablate the ratio between the size of the OOD fine-tuning data and the added amount of ID data in the multi-task setting in Fig. \ref{fig::ablationsteps}. While adding ID data leads to a decrease of the metrics for all amounts, the ratio of $0.25$ performs best and even decreases the OOD metrics. That underlines the importance of searching for the right amount of added ID data during multi-task fine-tuning.
Tab. \ref{tab:ablations_results_exiD} ablates the influence of learning goal and cost residuals, as well as the low-rank decomposition of Equ. (\ref{equ::low_rank_decomposition}). We also ablate if adding the residual on the output joint control trajectory is less effective than adding the residual on the cost. Our base model (Fine-tuning + LoRD) performs best underlining the importance of our design choices. 

\textbf{Parameters and Runtime}:
The additional LoRD introduces a small increase in parameters (nuPlan: 0.09\%, exiD: 0.42\%) and runtime (nuPlan train: 1.91\%, test: 1.33\%).

\section{Conclusion, Discussion, and Future Work}
This work presented an approach for adapting differentiable prediction, planning, and control. Our results showed improved performance on two datasets for structured policies in both open-loop and closed-loop settings. We also observed a significant gap between open-loop and closed-loop metrics, highlighting the need for more focus on closed-loop evaluation. Future work could explore merging LoRD weights with the base network. It should also investigate adapting to various shifts, such as new countries or weather, by adding specialized residuals or retraining a single residual for all OOD datasets, while addressing potential biases and ethical considerations in real-world deployment.

\addtolength{\textheight}{-6.2cm}   




\bibliographystyle{plain}
\bibliography{references}

\end{document}